%% file: main-paper.tex
\documentclass[conference]{IEEEtran}
\IEEEoverridecommandlockouts
\usepackage{cite}
\usepackage{amsmath,amssymb,amsfonts}
\usepackage{graphicx}
\usepackage{textcomp}
\usepackage{xcolor}
\usepackage{comment}
\usepackage{gensymb}
\usepackage{caption}
\usepackage[linesnumbered,lined,figure,vlined]{algorithm2e}
\usepackage{multicol}
\usepackage{url}
\usepackage{hanging}
\usepackage{diagbox}

\def\BibTeX{{\rm B\kern-.05em{\sc i\kern-.025em b}\kern-.08em
    T\kern-.1667em\lower.7ex\hbox{E}\kern-.125emX}}
\begin{document}

\title{{\fontsize{14}{16}\bfseries\selectfont A Method for the Runtime Validation of AI-based Environment Perception in Automated Driving Systems*}}
\author{
		\IEEEauthorblockN{
			Iqra Aslam\IEEEauthorrefmark{1},
			Abhishek Buragohain\IEEEauthorrefmark{2},
			Daniel Bamal\IEEEauthorrefmark{3},
			Adina Aniculaesei\IEEEauthorrefmark{4},
			Meng Zhang\IEEEauthorrefmark{5},
			and Andreas Rausch\IEEEauthorrefmark{6}}
		\IEEEauthorblockA{\IEEEauthorrefmark{0}
			Institute for Software and Systems Engineering, Technische Universit\"at Clausthal \\
			Clausthal-Zellerfeld, Germany\\
			Email: \{
			iqra.aslam\IEEEauthorrefmark{1},
			abhishek.buragohain\IEEEauthorrefmark{2},
			daniel.bamal\IEEEauthorrefmark{3},
			adina.aniculaesei\IEEEauthorrefmark{4},
			meng.zhang\IEEEauthorrefmark{5},
			andreas.rausch\IEEEauthorrefmark{6}
			\}
			@tu-clausthal.de}
	}
\maketitle

\begin{abstract}

Environment perception is a fundamental part of the dynamic driving task executed by Autonomous Driving Systems (ADS). Artificial Intelligence (AI)-based approaches have prevailed over classical techniques for realizing the environment perception. Current safety-relevant standards for automotive systems, International Organization for Standardization (ISO) 26262 and ISO 21448, assume the existence of comprehensive requirements specifications. These specifications serve as the basis on which the functionality of an automotive system can be rigorously tested and checked for compliance with safety regulations. However, AI-based perception systems do not have complete requirements specification. Instead, large datasets are used to train AI-based perception systems. This paper presents a function monitor for the functional runtime monitoring of a two-folded AI-based environment perception for ADS, based respectively on camera and LiDAR sensors. To evaluate the applicability of the function monitor, we conduct a qualitative scenario-based evaluation in a controlled laboratory environment using a model car. The evaluation results then are discussed to provide insights into the monitor's performance and its suitability for real-world applications. 

\textbf{\textit{Keywords-runtime monitoring; function monitor; dependable safety-critical system; automated driving system; perception system}}.
\end{abstract}

\input{introduction}
\input{related-work}
\input{concept}
\input{evaluation}
\input{summary}

\section*{Acknowledgment}
The authors express their gratitude for the collaboration of all cooperative partners. This work was carried out as part of the project \textit{safeWahr - Safe release and reliable series operation through continuous real-time monitoring of the environmental perception of autonomous vehicles} \cite{projectnameingerman:2021} under the grant number 19A21026E and was funded by the Federal Ministry for Economic Affairs and Climate Protection \cite{BMKW:online} in Germany.

\bibliographystyle{ieeetr} 
\bibliography{literature}

\end{document}

%% file: introduction.tex
\section{Introduction}
\label{sec:introduction}
In principle, fully autonomous vehicles are technically feasible. However, after the initial proof-of-concept testing under ideal conditions, e.g., in lab environments \cite{aniculaesei2023connected}, or on restricted test fields  \cite{aslam2023runtime}, further innovative verification and validation techniques are needed during the approval and release processes. These additional verification and validation phases are necessary to gather the required evidence for the safety and reliability of the autonomous vehicle in real-world scenarios. For the commercial approval of autonomous vehicles by certification bodies, the current state-of-the-art practices require verifying specific maneuvers using predefined test scenarios and statistically analyzing real-time data covering millions of kilometers of driving. For autonomous vehicles at Society of Automotive Engineers (SAE) Level 3/4 and above, the key challenge lies in ensuring the safe commercial release and the safe vehicle operation in all possible situations, not just only in those situations encountered during the system development, e.g., through random tests.

Today's automated driving systems are primarily designed to be fail-safe systems, capable of switching the ego-vehicle to a safe state, e.g., by activating an emergency brake. However, future ADSs must be designed as fail-operational systems, especially as there are many situations in which an immediate fail-state might not be readily accessible, e.g., when driving at high speed on the highway. Moreover, in case issues occur during the vehicle operation, the control over the dynamic driving task can no longer be simply handed back to the human driver, since human intervention is not mandated anymore at SAE L4+ \cite{SAEJ3016:2018}. Without a human fallback system, the ADS must be able to take over control and establish a safe state for the vehicle, if a problematic situation arises.

In recent years, a high-level functional architecture has been established for ADSs comprising three main subsystems: (1) environment and self-perception, (2) situation comprehension and action planning, and (3) trajectory planning and motion control \cite{Behere.2015}, \cite{Maurer.2015}. The environment and self-perception is particularly significant as it strongly impacts the performance of the entire ADS and the safety of the autonomous vehicle, as shown by a series of accidents involving (partially) automated driving functions, e.g., Tesla's autopilot. In the first notable accident in 2016, the Tesla's autopilot has failed to detect an articulated lorry driving in the opposite direction which was engaged in a turn maneuver, despite having been successfully tested over 200 million kilometers. In the context of a brightly lit sky, both the driver and the autopilot were unable to recognize the lorry, which had white sides \cite{Tesla:2016}. In response to this accident, Tesla announced the introduction of Shadow Mode to enhance the safety of its autopilot \cite{Golson:2016}. An approach for continuous monitoring of autonomous driving functions for development, validation and series operation has been proposed in in \cite{Mauritz:2014} and demonstrated for the lane changing functionality of the highway pilot in \cite{Mauritz:2016}. This approach essentially extends the concept of shadow mode, by addressing two questions: (1) does the autonomous driving function operate correctly (qualitative oracle) and (2) is the autonomous driving function currently operating in a known environment (quantitative oracle) \cite{Mauritz:2016}. 

There is a noticeable gap in research work regarding the validation of environment and self-perception methods applied in automated driving applications. AI-based approaches have prevailed over classical approaches for realizing environment perception, as the former are used both in image processing and in signal processing of other raw sensor data, such as radar. Current safety-relevant standards for automotive systems, e.g., ISO 26262 \cite{ISO26262:2018} and ISO 21448 \cite{ISO21448:2022}, assume the existence of complete requirements specifications. The system development process is usually organized using a structured process model, e.g., the V-model. However, challenges arise with AI-based perception systems since they do not have complete requirements specifications. Instead, the development usually begins with incomplete artifacts, e.g., system requirements formulated for the entire ADS structured test cases derived from other artifacts, e.g., in the format of OpenSCENARIO \cite{OpenSCENARIO:2022} or OpenDRIVE \cite{OpenDRIVE:2023} formats. Additionally, the development of AI-based systems requires  extensive training. For instance, for the development of an AI-based system for pedestrian detection an substantial training data set consisting of diverese pedestrian images is required.

The approach outlined in \cite{Mauritz:2016} by Mauritz and his co-authors can be understood as developing a dependability cage for monitoring the lane changing functionality of the highway pilot. While their focus is monitoring the entire autonomous driving function without particularly considering the environment perception, in this work we focus on monitoring the environment and self-perception subsystem of an ADS. We propose a dependability cage for validating an environment and self-perception system including redundant perception components that operate with multiple sensor data sources. In this dependability cage, an essential component is a function monitor which evaluates the consistency of the outputs produced by the redundant perception components during the ADS operation. This function monitor is derived from a high-level safety requirement established for the environment and self-perception subsystem. In this work, we evaluate the dependability cage approach for AI-based environment perception qualitatively in a lab environment using predefined test scenarios.

The rest of the paper is structured as follows. Section \ref{sec:related-work} gives an overview of related work in the area of verification and validation of AI-based system, with a specific focus on environment perception systems in both robotic and automotive applications. Section \ref{sec:concept} presents the dependability cage approach for the monitoring and validating AI-based environment perception during the ADS operation. In Section \ref{sec:evaluation}, we conduct a a qualitative scenario-based evaluation and discuss the obtained results. Section \ref{sec:summary} concludes the paper by summarinzing its contributions and outlining potential future work directions.

%% file: related-work.tex
\section{Related Work}
\label{sec:related-work}
Environment- and self-perception is an integral part of the dynamic driving task that is carried out by the ADS. Primarily based on AI models, it provides essential inputs for further safety-critical functions of the ADS, such as Situation Comprehension and Trajectory Planing. Through the interpretation of various raw sensor data into detailed semantic information about the surrounding driving environment, the AI-based environment perception subsystem enables the ADS to complete its decision making, motion planning and control command execution, by relying on its respective planning and decision components. 

For this reason, monitoring and evaluating the functional behavior and the performance of AI-based environment perception systems at runtime is extremely important for the safety of the autonomous vehicle. Recently, diverse research approaches have addressed the safety issues of AI-based environment perception. Czarnecki \cite{czarnecki2018towards} identifies a set of influencing factors for AI-based environment perception: (1) conceptual uncertainty, (2) development situation and scenario coverage, (3) situation or scenario uncertainty, (4) sensor properties, (5) labeling uncertainty, (6) model uncertainty, and (7) operational domain uncertainty. Identifying these factors is understood as the first step, which should be followed by a systematic analysis of their impact on the perceptual uncertainty. In addition, methods to eliminate or reduce their negative effects on the perceptual uncertainty. Subsequently, mitigation measures are proposed to be applied for the cases when the control of the negative effects is not possible \cite{czarnecki2018towards}. The concept in \cite{czarnecki2018towards} revolves around using these methods to gather the essential evidence to substantiate claims regarding the environment perception uncertainty in safety cases that contribute to demonstrating the safety of the overall autonomous vehicle \cite{czarnecki2018towards}.


Another survey presented in \cite{rahman2021run} identifies various research directions concerning the runtime performance monitoring of AI-based perception in autonomous robots. One direction encompasses approaches using past examples of failures and successes or similarity of operational context to previous experiences to predict the quality of perception output \cite{rahman2021run}. The second direction involves by methods that detect inconsistencies in perception output, using the input data from a single sensor or from multiple sensors, or by comparing outputs from different models \cite{rahman2021run}. The third direction focuses on methods for uncertainty estimation by indicating low confidence in prediction output, calculating confidence scores as a measure of the target model's output quality, and detecting anomalies \cite{rahman2021run}. Several other studies also aim to validate the accuracy or robustness of the perception system outputs concerning specific inputs of the neural network. However, the approaches surveyed in \cite{xiang2018verification} and \cite{leofante2018automated} are primarily limited to offline verification.


In addition to estimating uncertainty, some recent studies have shifted their primary focus to object detection as the main task of the perception system under analysis. Thus, Feng et al. \cite{feng2021labels} evaluate LiDAR-based object detectors using the Jaccard Intersection over Union (IoU) metric and KITTI and Waymo datasets, incorporating label uncertainty specifically for the bounding boxes of a particular object class, e.g., the object class \textit{car}.  In \cite{rahman2021per}, the authors propose a framework to predict performance monitoring of object detection at runtime without relying on any ground-truth data. Meanwhile, in \cite{rahman2021online}, they monitor the performance of the object detector deployed on mobile robots by predicting the quality of its mean average precision using a sliding window technique over the input frames. However, the approach presented in \cite{gupta2020online} for monitoring of neural networks that estimate 3D human shapes and poses from images is limited to human pose estimation. 

In previous research, the majority of studies have made significant contributions towards the evaluation methods, applications and understanding of uncertainty in perception systems. Nevertheless, most of these studies focus on a single object detection network. They do not address the comprehensive validation of perception system outputs concerning reference sensors or the comparative analysis of outputs from redundant perception systems. Additionally, there is a research gap in detecting perception failures or incorrect behaviors of environment perception systems at runtime to ensure the safety of the perception-equipped system. In this work, we propose an approach for the runtime validation of environment perception in autonomous vehicles (AVs), by analysing and comparing the outputs of two redundant perception systems utilizing camera and respectively LiDAR sensors data. We evaluate this approach qualitatively using predefined scenarios and a model car in a lab environment. 

%

%% file: concept.tex
\section{Integrated Safety Architecture with Dependability Cage for the Runtime Validation of AI-based Environment Perception in Automated Driving Systems}
\label{sec:concept}
%
\begin{figure}
    \centering
    \includegraphics[width=\columnwidth, keepaspectratio]{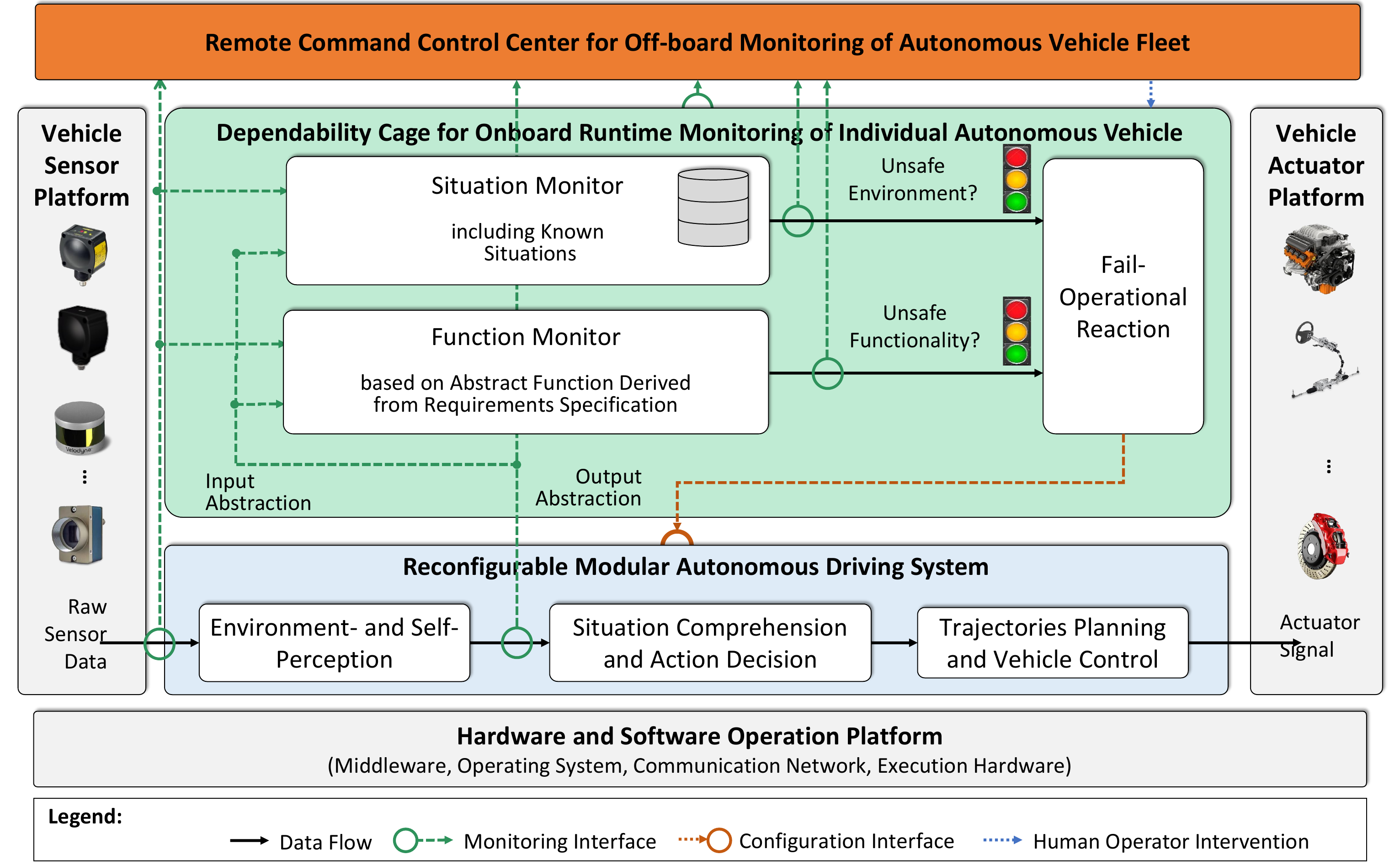}
    \caption{High-level Architecture of the Dependability Cage for Runtime Validation AI-based Environment Perception Systems.}
    \label{fig:high-level-architecture-function-situation-monitor}
    \vspace{-5mm}
\end{figure}
This section introduces the Dependability Cage approach for the runtime monitoring and validation of AI-based environment perception in automated driving systems. Figure \ref{fig:high-level-architecture-function-situation-monitor} illustrates the high-level architecture of this dependability cage, that can be understood as an instantiation of the overarching concept of the Dependability Cage approach, presented first in \cite{Aniculaesei.2018}. The subsequent sections offer a detailed introduction to the architecture of the function monitor, derived on the basis of the Dependability Cage approach.

\subsection{Overall Concept of the Dependability Cage Approach}
\label{subsec:concept-overall-idea-dependability-cage}
In the initial position paper, the Dependability Cage approach was introduced to address three main challenges in the development of autonomous systems: (1) guaranteeing the safe behavior of an autonomous system in an unknown and uncertain environment, (2) ensuring the safe behavior for all safety-critical system components, including machine-learning based components, even when deviations are detected in their behavior during system operation time, and (3) guaranteeing and improving the relevance and completeness of test cases for the validation of the system under test \cite{Aniculaesei.2018}.

To tackle these challenges, the paper in \cite{Aniculaesei.2018} proposes an iterative development process consisting of three primary stages. The first stage is \textit{Dependability Cages Engineering and Training in System Development}, in which the dependability cages are engineered in parallel to the autonomous system and tested using simulation tests or tests in a restricted and controlled lab environment \cite{Aniculaesei.2018}. In the second stage, \textit{Runtime System Observation and Resilience System Stabilization}, the dependability cages are used to monitor the system behavior during its operation and record any deviation in the system behavior  compared to the test results obtained during system development as well as any novel situations that occur in the environment \cite{Aniculaesei.2018}. In the third stage, \textit{Monitored Data Analysis and Goal-Oriented System Evolution for Dependability Improvement}, the observations logged during system operation are leveraged to improve the development artifacts during the subsequent system development cycle \cite{Aniculaesei.2018}. The deployment of the dependability cages on the actual system during its operation is facilitated by a modular platform architecture used for the seamless development and operation of the system, monitor and system environment \cite{Aniculaesei.2018}.
The end goal of this iterative development process is to continuously improve the system's quality in terms of its dependability requirements. For further details on the development process and its phases, the reader is referred to \cite{Aniculaesei.2018}.
\begin{figure*}
    \centering
    \includegraphics[width=.7\textwidth, keepaspectratio]{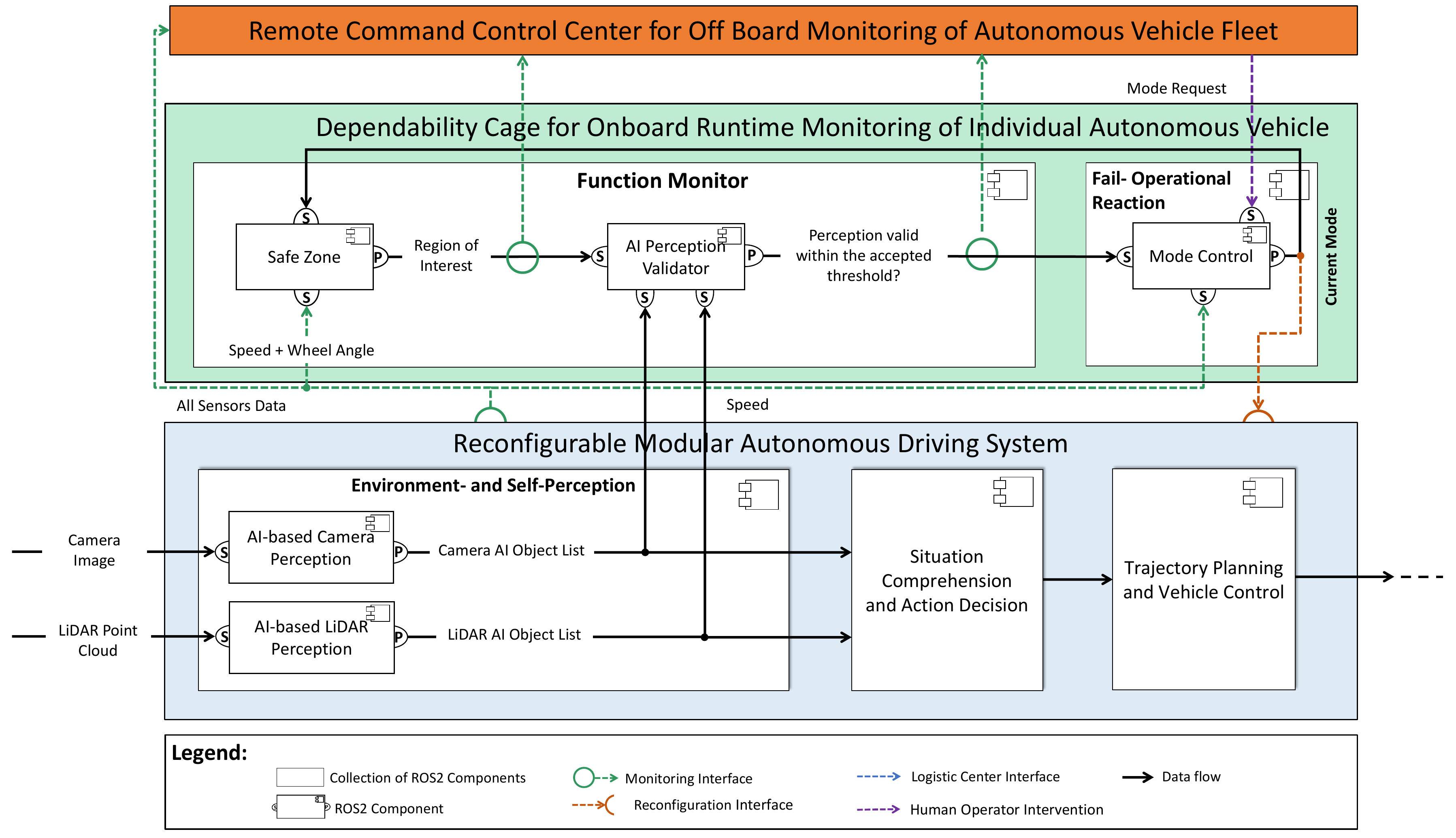}
    \caption{Architecture of the Function Monitor for AI-based Perception Systems.}
    \label{fig:function-monitor-environment-perception}
    \vspace{-5mm}
\end{figure*} 
The safety architecture proposed for autonomous systems in \cite{Aniculaesei.2018} draws its inspiration from the second phase of the iterative development process, \textit{Runtime System Observation and Resilience System Stabilization}. This phase is carried out through a continuous monitoring framework that observes the behavior of the overall system, as shown in \cite{Aniculaesei.2018}, \cite{Felix.2022}, and \cite{aslam2023runtime}. In this paper, we refine the focus of this monitoring framework and tailoring it to analyze the environment perception subsystem of an ADS, rather then the ADS as a whole. The monitoring framework consists of two types of monitors, a \textit{function monitor} and a \textit{situation monitor}. Additionally, it involves a component responsible for defining the fail-operational reaction of the ADS based on the results of these two runtime monitors, as depicted in Figure \ref{fig:high-level-architecture-function-situation-monitor}. 

The responsibility of the \textit{situation monitor} - denoted as the quantitative monitor in the original position paper \cite{Aniculaesei.2018} - is to evaluate the input abstract situations used in the environment and self-perception subsystem of the ADS. During system operation, the situation monitor assesses if the input situations encountered by the ADS align with those considered during the development phase of the environment perception. As the research in this paper does not focus on the situation monitor, the reader is referred to \cite{Rausch.2021} and \cite{habib2023} for a more detailed description of its concept and functionality.
    
On the other hand, the \textit{function monitor} - denoted as the qualitative monitor in the original position paper \cite{Aniculaesei.2018} - evaluates if there is a critical deviation in the behavior of the environment and self-perception subsystem during the operation of the ADS. The function monitor consists of an abstract behaviour’s boundary function and a conformity oracle. The abstract behaviour’s boundary function dynamically computes a set of safety boundaries, conceptualized as a Region Of Interest (ROI) for the ADS. The conformity oracle assesses the abstraction of the environment and self-perception's outputs in order to check if these outputs are consistent with each other within the safety boundary. Consistency is evaluated based on certain threshold values that are established by empirical tests during the development phase of the environment and self-perception of the ADS. The architecture of the function monitor is explained in detail in the Section \ref{subsec:concept-architecture-dependability-cage}.

\subsection{Dependability Cage for AI-based Environment Perception in the ADS's Integrated Safety Architecture}   
\label{subsec:concept-architecture-dependability-cage}
The dependability cage for AI-based environment perception is integrated in the three layered safety architecture developed for ADSs. Beginning from the top, the first layer is represented by the \textit{Remote Command Control Center (CCC)}, where the sensors data stream is visualized along with the results provided by the components in the layers below \cite{Felix.2022}, \cite{aniculaesei2023connected}. The remote CCC has been previously showcased in a separate work \cite{aniculaesei2023connected} and is therefore not the primary focus of the current paper. 

The research focus of this paper lies in the middle layer and the bottom layer of the integrated safety architecture. The middle layer encompasses the dependability cage for the AI-based environment perception of the ADS. The dependability cage consists of two main components: (1) the function monitor, responsible for observing and analyzing the environment perception system during the ADS's operation and (2) the fail-operational reaction component, which triggers a fail-operational reaction of the ADS based on the results of the function monitor. The third layer represents the re-configurable modular autonomous driving system \cite{everding2023dynamically}, which draws inspiration from previous work \cite{Behere.2015}, \cite{Maurer.2015}, and \cite{Behere.2016}, and consists of three main subsystems: (1) Environment and Self-Perception, (2) Situation Comprehension and Action Decision, and (3) Trajectory Planning and Vehicle Control.

The environment and self-perception subsystem consists of two components, AI-based Camera Perception and AI-based LiDAR Perception. These components use camera and respectively LiDAR sensor data to detect objects in the AV's environment. They implement safety-critical machine-learned functions for the ADS's operation. Each component produces an object list, denoted as Camera AI-based (CAI) object list and LiDAR AI-based (LAI) object list. The object lists are used by the other components in the architectural pipeline of ADS. Several pieces of information are provided for each object in the two object lists: (1) object's class, (2) the object's dimensions, height and width, (3) object's distance from the ego-vehicle, (4) sensing timestamp, and (5) confidence level of detection. Figure \ref{fig:function-monitor-environment-perception} provides an overview of the integrated safety architecture with a focus on the function monitor and the environment and self-perception system.    

The function monitor observes the behavior of the environment and self-perception subsystem against to a specified safety requirement. The safety requirement is informally formulated in a controlled natural language as follows:

\medskip
\begin{hangparas}{.50in}{1}
	\textbf{Safety Requirement (Informal Specification).} The environment and self-perception system must always consistently detect the objects located inside the autonomous vehicle's region of interest using at least two different sensor data sources.
\end{hangparas}
\medskip

This safety requirement mandates two aspects: first, that a ROI is computed around the AV, and second, that the objects detected by the perception components in the vehicle's region of interest are consistent with each other. The function monitor consists of two components, which allow it to check this safety requirement during the ADS's operation, denoted as \textit{Safe Zone} and \textit{AI Perception Validator}.
 
\subsubsection{Safe Zone} This component utilizes various parameters of the AV, e.g., current speed, steering angle, physical dimensions of the AV, acceleration and deceleration, to dynamically calculate ROI. The ROI expands around the AV in the vehicle's direction of travel, and is divided in two areas: a \textit{focus zone} marked in orange and a \textit{clear zone} marked in green around the ego-vehicle. Figure \ref{fig:model_car} gives a visual intuition of the ROI, depicted on the Graphical User Interface (GUI) of the remote CCC. The ROI around the ego-vehicle is understood as a safety-critical area in which the outputs of the two perception components align consistently. Further details regarding the algorithm used for the ROI computation can be found in \cite{Felix.2022}.
   
\subsubsection{AI Perception Validator} 
\begin{figure}
    \centering
    \includegraphics[width=\columnwidth, keepaspectratio]{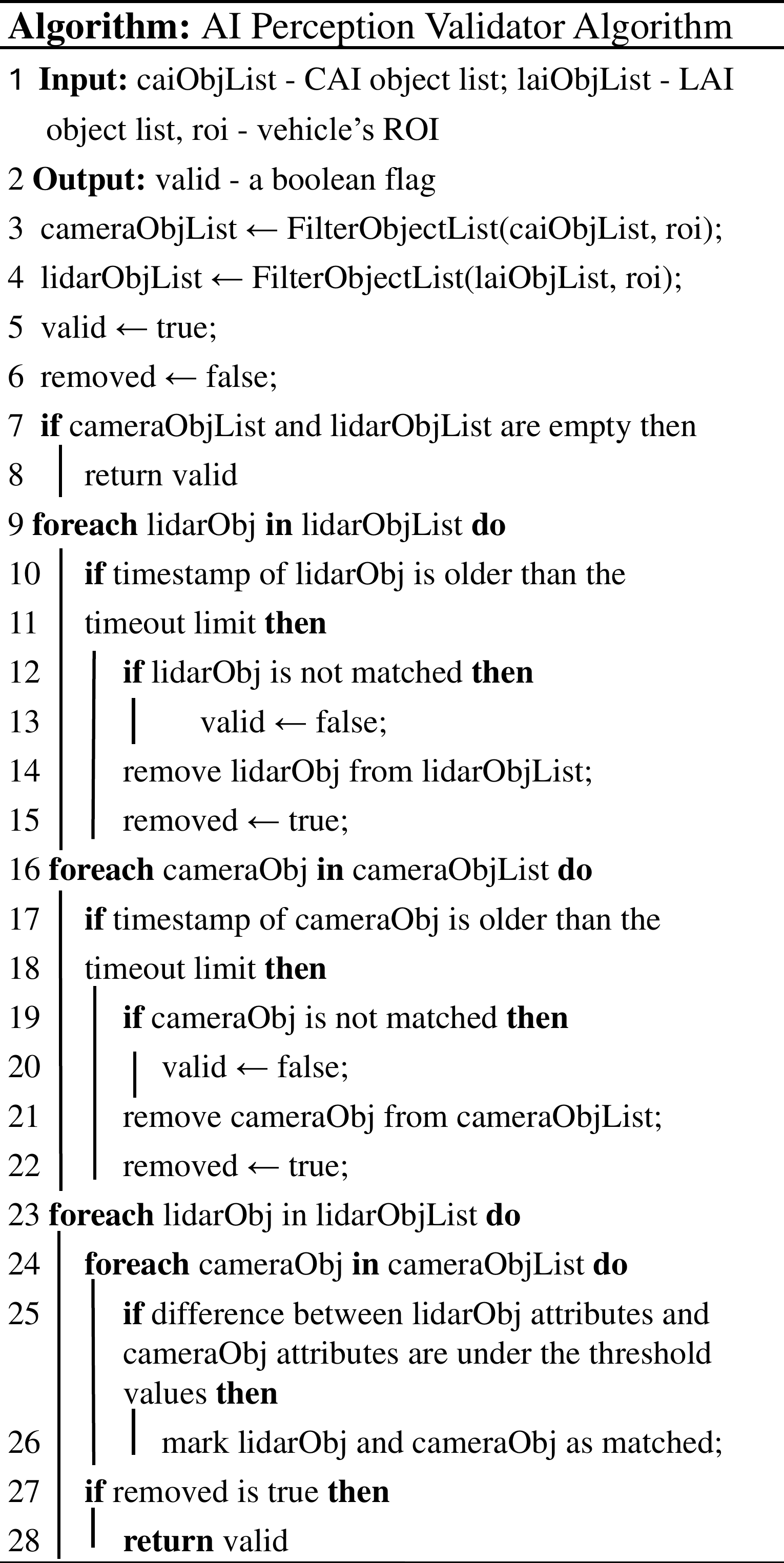}
    \caption{AI Perception Validator Algorithm.}
    \label{fig:algorithm}
    \vspace{-5mm}
\end{figure}

This component takes as inputs the ROI computed by the Safe Zone component and the object lists produced by the LiDAR-based and camera-based perception components and computes a boolean flag \textit{valid}, which indicates if the object lists from both perception components are consistent within certain threshold limits (as indicated in lines 1 - 2 in Figure \ref{fig:algorithm}). The AI Perception Validator leverages the computed ROI to prune the set of objects detected by the camera and LiDAR sensors in the respective fields of view (as indicated in lines 3 - 4 in Figure \ref{fig:algorithm}). It prioritizes those objects detected in close proximity to the ego-vehicle, i.e. inside the ROI of the vehicle. Such a prioritisation differentiates clearly between the objects situated inside the ROI that are safety-relevant for the AV, from the objects situated outside the ROI, which do not pose an immediate safety concern. The threshold limits for determining the consistency of the two object lists are established through a Hazard Analysis and Risk Assessment (HARA). Given that the camera and LiDAR sensors operate at different time rates due to their inherent configuration \cite{rozsa2023virtually}, the CAI object list and the LAI object list will be generated at different frequencies. This time synchronization problem is addressed by using a timeout limit in the AI Perception Validator for the timestamp of the detected objects. It means that an object with a timestamp older than the timeout limit is filtered out from the respective object list (as indicated in lines 9 - 22 in Figure \ref{fig:algorithm}). Subsequently, the AI Perception Validator compares the attributes of each object in the CAI object list with the corresponding attributes of each object in the LAI object list, e.g. object class, object distance from the ego vehicle, width and height of the object. If the difference between the respective attributes does not exceed the respective threshold value determined through the HARA analysis, then it is considered a match between both object lists and the AI Perception Validator returns $\mathit{true}$ (as indicated in lines 23 - 26 in Figure \ref{fig:algorithm}). It means that the outputs of the two perception components are consistent with each other and all objects have passed the validation and matching criteria. Another case for returning $\mathit{true}$ is that both CAI object list and LAI object list are empty. In this case, there is no need to compare the two object lists. If at least one object in any of the two lists does not meet the comparison criteria, the AI Perception Validator returns $\mathit{false}$, meaning that the outputs of the two perception components are inconsistent (as indicated in lines 27 - 28 in Algorithm \ref{fig:algorithm}). 
\vspace{5mm}
The result of the AI Perception Validator as well as the two object lists are forwarded to the remote CCC for visualization (see Section \ref{sec:evaluation}). Additionally, the result computed by the AI Perception Validator serves as an input in the component \textit{Mode Control}. In case of inconsistency between the two object lists, this component is responsible for triggering a fail-operational reaction of the AV, by gracefully degrading the ADS functionality. This process is similar to approach outlined in \cite{Felix.2022}. However, since the primary research focus of this paper is the function monitor, the definition, implementation and evaluation of the corresponding fail-operational reaction will be addressed in future work.

%

%% file: evaluation.tex
\section{Evaluation and Discussion of Results}
\label{sec:evaluation}
This section presents the evaluation of the function monitor for the runtime validation of AI-based environment perception in ADS. The first part of this section introduces a detailed overview of the setup, including both hardware and software details (see Section \ref{subsec:evaluation-setup}). Subsequently, in Section \ref{subsec:evaluation-scenarios}, various test scenarios and several working hypotheses are defined. A qualitative scenario-based evaluation is conducted to assess the defined hypotheses and the obtained results are discussed.

\subsection{Evaluation Setup}
\label{subsec:evaluation-setup}
\subsubsection{Physical Hardware Platform and Test Track}

\begin{figure}
    \centering
    \includegraphics[width=0.95\columnwidth, keepaspectratio]{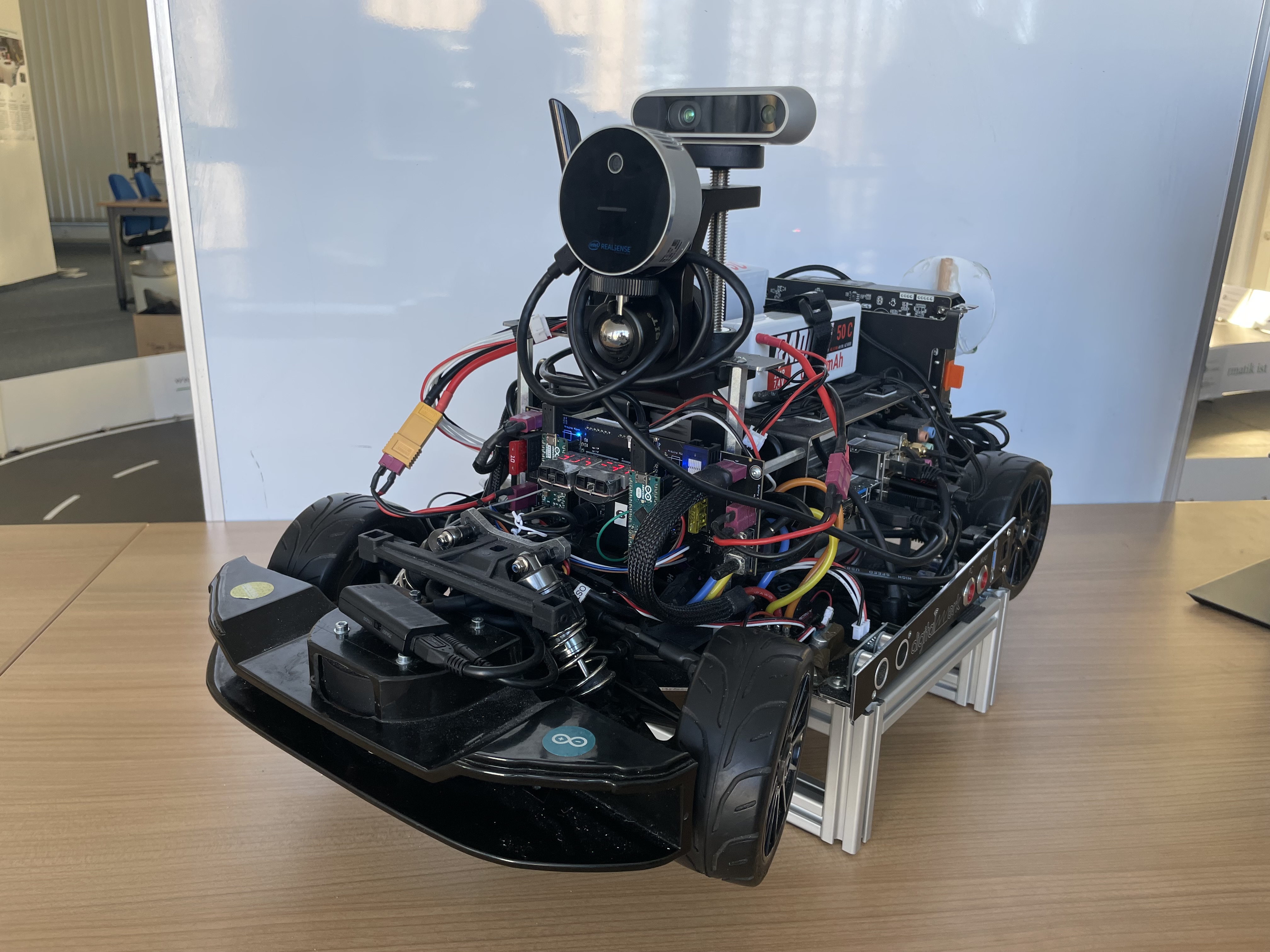}
    \caption{Model Car}
    \label{fig:model_car_lab}
    \vspace{-5mm}
\end{figure}

For the evaluation of the function moniotr concept, a model vehicle on the scale of 1:8, developed by Digitalwerk \cite{Welcomet61:online}, is chosen as the physical hardware platform. The model vehicle is equipped with a wide range of sensors, including a mono camera, a LiDAR sensor, wheel speed sensors, ultrasonic sensors, a Global Positioning System (GPS) sensor, and an Inertial Measurement Unit (IMU). 

To enhance its environmental perception capabilities for the validation of the function monitor, further sensors have been installed on the model vehicle, e.g., an Intel RealSense LiDAR camera (L515)\cite{LiDARCam3:online} and a stereo vision camera (D435f) \cite{DepthCam5:online}. The LiDAR camera provides sensor input data for the LiDAR AI-based perception component. Although a high-resolution 3D LiDAR sensor would have been ideal, the model vehicle's limited power supply led to the deployment of a LiDAR camera with lower power requirements as a good compromise solution, which still provides adequate data output. Both sensors have been calibrated based on the vehicle's rear axle, to ensure that generated object lists are in the same coordinate system, specifically in the vehicle coordinate system. This alignment is crucial for an accurate and coherent comparison of redundant perception systems. Figure \ref{fig:model_car_lab} illustrates the model car with the LiDAR camera and stereo vision camera mounted on it.

\begin{figure}
    \centering
    \includegraphics[width=0.92
    \columnwidth, keepaspectratio]{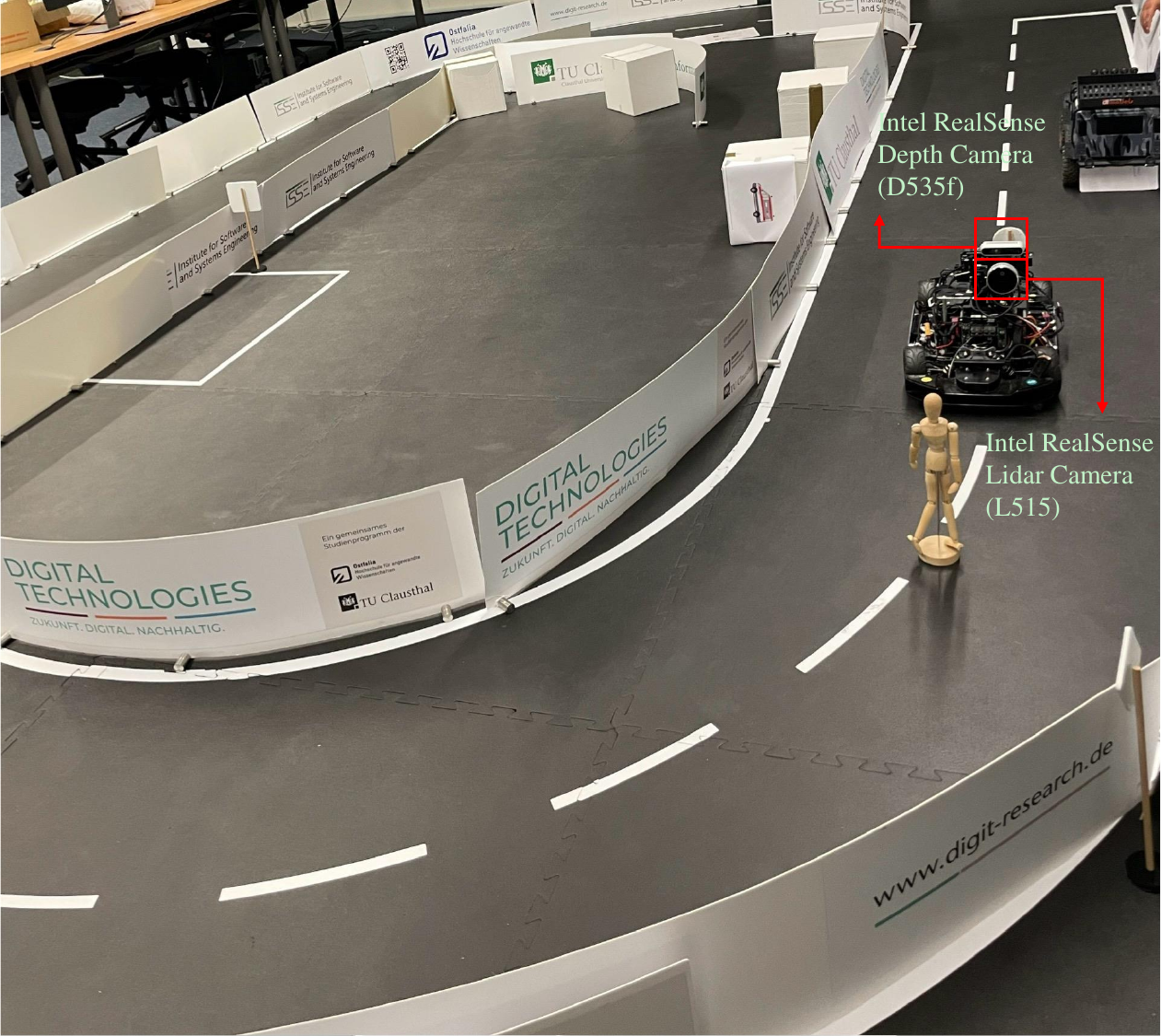}
    \caption{Model Car: bird's-eye view in a Lab Environment.}
    \label{fig:model_car}
    \vspace{-5mm}
\end{figure}

The test track utilized for the evaluation was constructed in the lab environment using modular martial arts mats. Each black mat measures $1 \text{m} \times 1 \text{m}$ and is adorned with street markings and track walls \cite{warnecke2020teaching}. Figure \ref{fig:model_car} depicts the model vehicle placed on the test track along with other objects that emulate other traffic participants, such as a pedestrian represented by a wooden human dummy, and elements of the road infrastructure, e.g., traffic light. 
\subsubsection{Implementation Details}
The function monitor conducts a consistency comparison between two object lists, a CAI object list and a LAI object list. These two lists are generated by respective AI-based perception components, one based on camera input and the other on LiDAR input. Both perception components apply YOLO Nano 2D object detectors \cite{wong2019yolo}, which yield 2D bounding boxes with object class names and and their respective confidence scores, but lack the distance information between the ego-vehicle and the corresponding objects. By leveraging the LiDAR point cloud provided by the LiDAR camera, we computed the distance between the model vehicle and the objects, referred to as depth, thereby producing 2.5D bounding boxes. The 2.5D bounding boxes differ from the 3D bounding boxes in that the latter include all three dimensions, i.e., height, width and length of the bounding box. 

The outputs of the environment and self-perception subsystem along with the result of the function monitor are visualized in the GUI of the remote CCC (see Section \ref{sec:concept}). Figure \ref{fig:person_inside_sz} depicts the visualization of the function monitor result in the Car Selection panel of the remote CCC, utilizing a flag called \textit{AI perception Validator}. The flag's color indicates different results of the function monitor: (1) \textcolor{green}{\textbf{green}} - denotes consistency between the two object lists, (2) \textcolor{red}{\textbf{red}} - signifies inconsistency, and (3) \textcolor{black}{\textbf{black}} - denotes data not being received by the AI Perception Validator component in the function monitor. 

In the center of Figure \ref{fig:person_inside_sz}, two panels display the view of the LiDAR camera (upper panel) and the stereo vision camera (lower panel) with the bounding boxes corresponding to each object list highlighted in green on their respective sensor view. Additionally, the Sensor Visualization panel depicts the ROI computed around the model vehicle and comprising a focus zone, marked in orange, and a clear zone, marked in green, as introduced in Section \ref{subsec:concept-architecture-dependability-cage}. 
The implementation of each component is based on the decentralized middleware ROS2, facilitating the communication between the ROS2 components through the publish-subscribe pattern. This solution provides advantageous features such as self-adaptation and component reconfiguration at runtime, aligning with the distributed nature of the AV, where components are distributed across different electronic control units (ECUs) \cite{raulf2022dynamic}. Furthermore, the real-time capabilities of ROS2 make it appropriate for ensuring the safety and reliability of the ADS.

\subsection{Definition of Test Scenarios and Research Hypotheses}
\label{subsec:evaluation-scenarios}
To evaluate the function monitor concept, we conducted a qualitative scenario-based assessment. We defined several test scenarios along with working hypotheses to guide the evaluation. The test scenarios range from simple to complex, starting with a single static object and gradually increasing the scenario complexity, by incorporating multiple static objects in a static environment. Each test scenario includes a description of the physical actions of the model vehicle and its environment. Following three test scenarios were defined evaluating the function monitor:

\begin{hangparas}{.50in}{1}
	\textbf{Test Scenario 1 (TS 1).} The model vehicle is stationary on the test track and a pedestrian (represented by a wooden human dummy) is placed in front of the model vehicle, outside of its ROI. The pedestrian is placed in such a way that it is detected by the LiDAR camera, but not by the stereo vision camera.
\end{hangparas}
\begin{hangparas}{.50in}{1}
	\textbf{Test Scenario 2 (TS 2).} The model vehicle is stationary on the test track and a pedestrian (represented by a wooden human dummy) is placed in front of the model vehicle, inside of its ROI. The pedestrian is positioned in such a way that it is detected by the LiDAR camera, but not by the stereo vision camera.
\end{hangparas}
\begin{hangparas}{.50in}{1}
	\textbf{Test Scenario 3 (TS 3).} The model vehicle is stationary on the test track and a traffic light is placed in front of the model vehicle, inside of its ROI. The traffic light is positioned so that both the LiDAR camera and the stereo vision camera are able to detect it.
\end{hangparas}
In addition to the test scenarion, several research hypotheses are formulated in this paper to assess the expected performance of the function monitor. The function monitor has been evaluated in the defined test scenarios with respect to the following two hypotheses:

\begin{hangparas}{.50in}{1}
	\textbf{Hypothesis 1 (H1).} The function monitor accurately identifies that the CAI object list and the LAI object list are consistent with each other.
\end{hangparas}

\begin{hangparas}{.50in}{1}
	\textbf{Hypothesis 2 (H2).} The function monitor accurately identifies that the CAI object list and the LAI object list are not consistent with each other.
\end{hangparas}

\subsection{Discussion of Results}
\label{subsec:evaluation-results-discussion}
\begin{figure}
    \centering
    \includegraphics[width=\columnwidth, keepaspectratio]{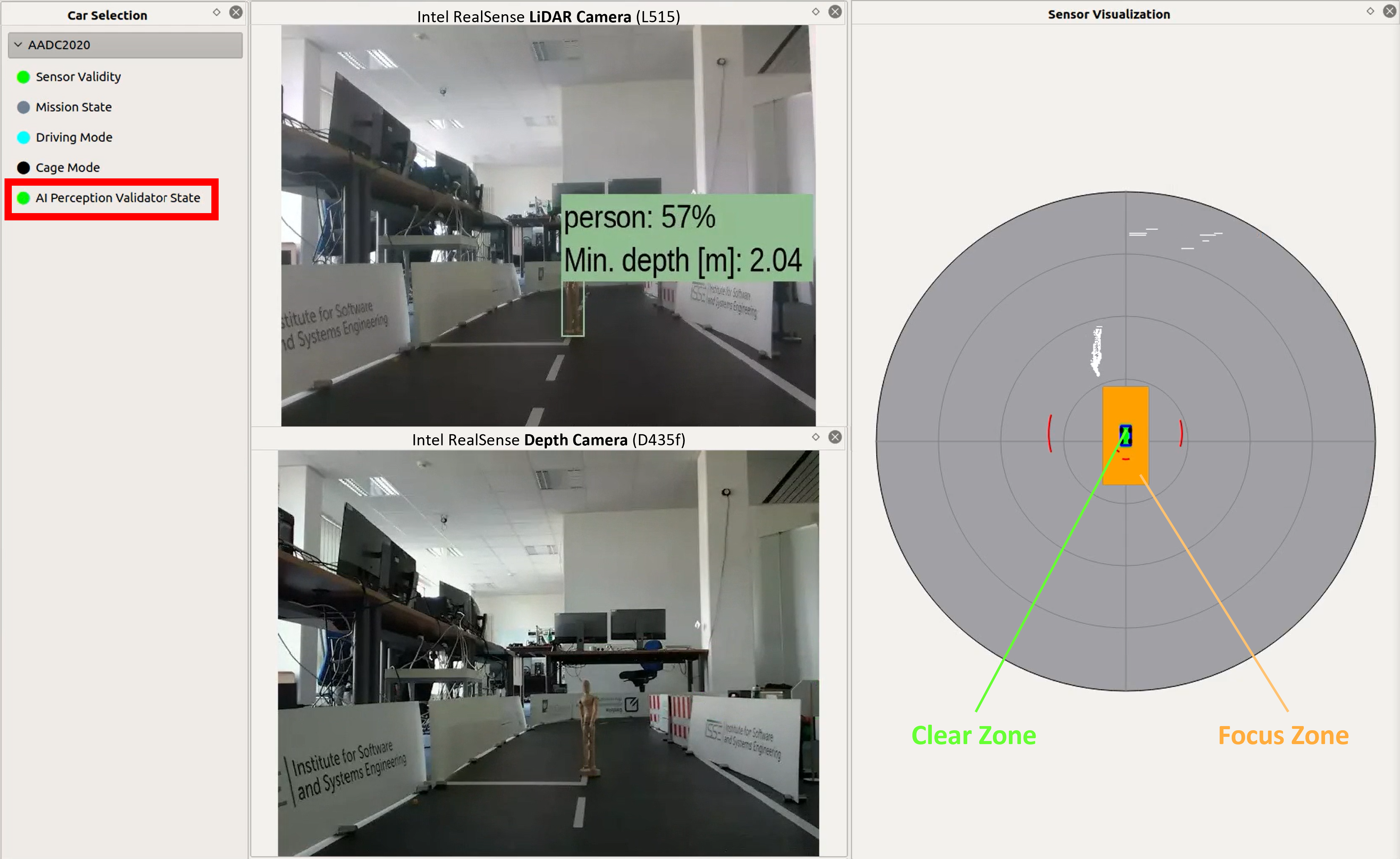}
    \caption{Wooden Human Dummy outside Model Vehicle's ROI.}
    \label{fig:person_outside_sz}
    \vspace{-5mm}
\end{figure}

The evaluation results of the function monitor on hypotheses H1 and H2 in all the test scenarios defined for the evaluation are presented in Table \ref{tab:evaluation-results}. In TS 1, in which the wooden dummy is placed in front of the vehicle and outside of its ROI, the LiDAR camera can detect it but the stereo vision camera cannot. In this scenario, the AI Perception Validator gives the result \textit{consistent} since the wooden dummy is outside of the model vehicle's ROI, and thus, both processed object lists are empty. Therefore, in TS 1, hypothesis H1 is \textit{true} and hypothesis H2 is \textit{false}. The CAI object list and the LAI object list along with the flag of the AI Perception Validator are visually depicted in Figure \ref{fig:person_outside_sz}, in which the AI Perception Validator flag shows a green status, indicating ``consistent object lists''.

\begin{table}[htpb]
    \caption{EVALUATION RESULTS OF THE FUNCTION MONITOR.}
    \label{tab:evaluation-results}
    \centering
    \begin{tabular}{|c|c|c|c|}\hline
    \diagbox[width=15em]{Hypotheses}{Test\\ Scenarios} & TS 1 & TS 2 & TS3 \\ \hline
      H1 & True & False & True \\ \hline
      H2 & False & True & False \\ \hline
    \end{tabular}
\end{table}

\begin{figure}   
    \centering
    \includegraphics[width=\columnwidth, keepaspectratio]{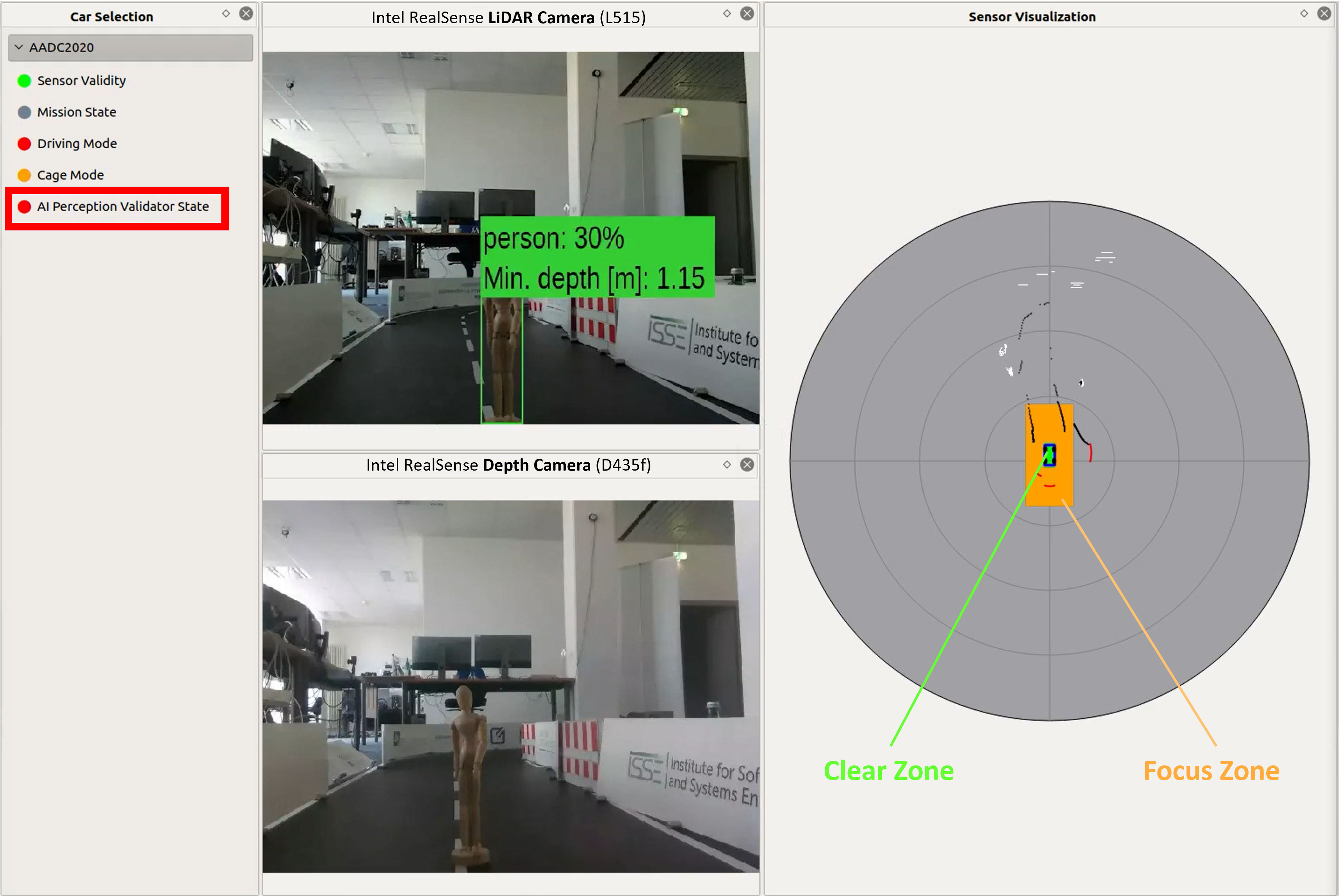}
    \caption{Wooden Human Dummy inside Model Vehicle's ROI.}
    \label{fig:person_inside_sz}
\end{figure}

\begin{figure}
    \centering
    \includegraphics[width=\columnwidth, keepaspectratio]{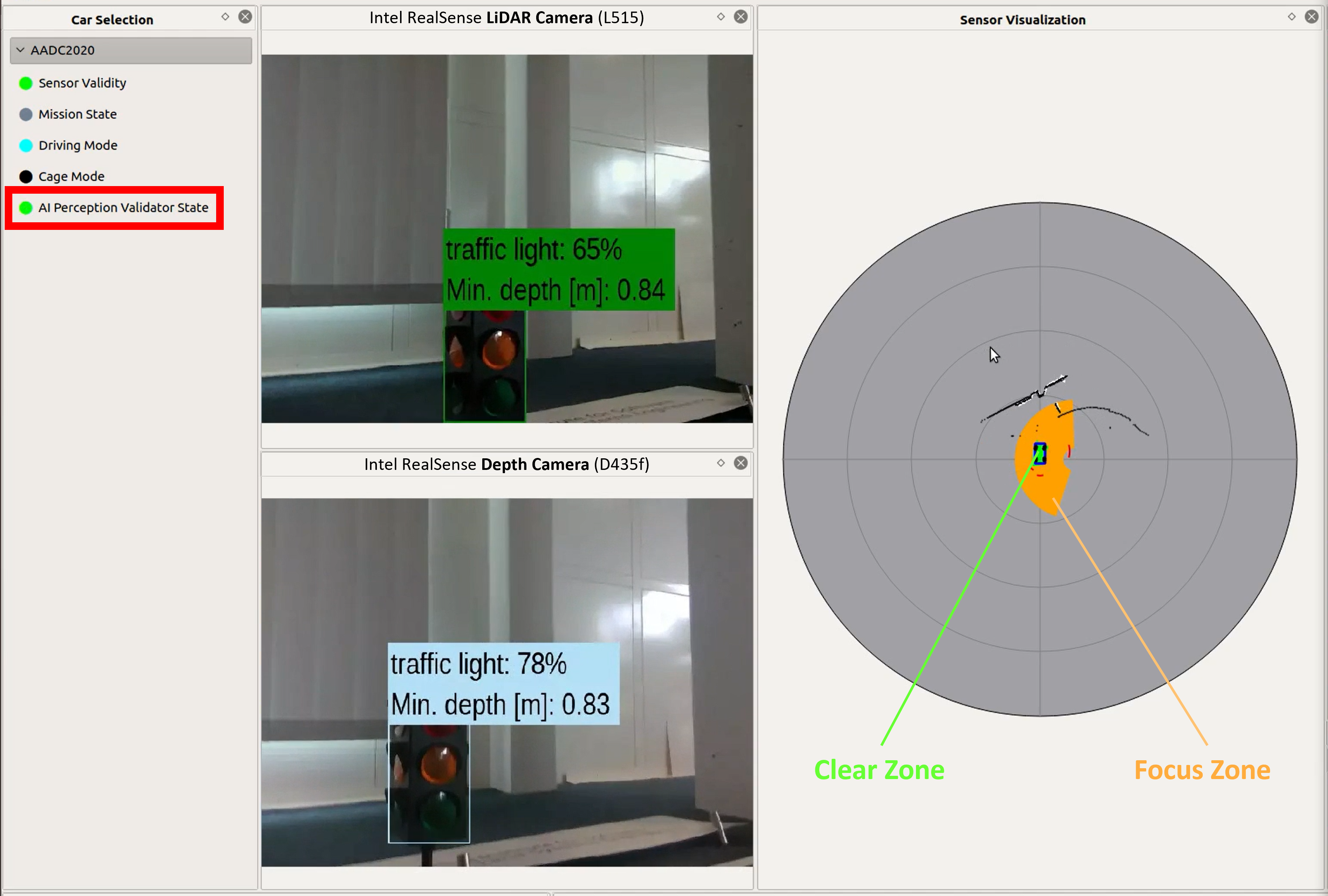}
    \caption{Traffic Light situated Inside Model Vehicle's ROI.}
    \label{fig:traffic_light}
    \vspace{-5mm}
\end{figure}

In TS 2, the wooden human dummy is placed again in front of the vehicle, but this time inside its ROI. The human dummy is placed so that it is detected by the LiDAR camera but not by the stereo vision camera. In this scenario, the AI Perception Validator returns \textit{inconsistent}, since there is at least an inconsistent object in either the CAI object list or the LAI object list, which in this case is the human dummy. Thus, in TS 2, hypothesis H1 is \textit{false} and hypothesis H2 is \textit{true}. The CAI object list and the LAI object list along with the flag of the AI Perception Validator are shown in Figure \ref{fig:person_inside_sz}. The AI Perception Validator flag shows a red status, indicating ``inconsistent object lists'' since the objects are inside the vehicle's ROI but not aligned with each other.
In TS 3, a traffic light is positioned in front of the model vehicle inside its ROI, so that both the LiDAR camera and the stereo vision camera can detect it. In this scenario, the AI Perception Validator returns \textit{consistent}, since the objects are inside the ROI and were detected by both sensors. Thus, in TS 3, hypothesis H1 is \textit{true} and hypothesis H2 is \textit{false}. The CAI object list and the LAI object list along with the flag of the AI Perception Validator are shown in Figure \ref{fig:traffic_light}, in which the AI Perception Validator flag indicates a green status, indicating ``consistent object lists''.

The results obtained in the three test scenarios have confirmed the formulated hypotheses in the defined test scenarios, showing that the function monitor is operating as intended.
%

%% file: summary.tex
\section{Conclusion}
\label{sec:summary}

This paper outlines a method for the runtime monitoring and validation of AI-based environment perception systems employed in autonomous driving contexts. It builds upon the Dependability Cage approach, initially proposed in \cite{Aniculaesei.2018}, with a specific focus on the environment and self-perception subsystem in an ADS. The environment perception consists of two redundant perception components tasked with object detection in the ego-vehicle surrounding environment, which leverage multiple sensor data sources, e.g., LiDAR and camera. The dependability cage for the environment perception comprises a function monitor and a fail-operational reaction component. The function monitor checks at runtime whether the outputs of the two perception components remain consistent. Meanwhile, the fail-operational reaction component dictates the fail-safe or the fail-operational reaction of the ADS based on the feedback of the function monitor. This study was primarily focused on the function monitor, which was evaluated qualitatively using predefined test scenarios and a model car in a lab environment. The results of the evaluation demonstrated that the function monitor works as expected. 

The test scenarios employed in the evaluation focused on relatively simple driving situations, with stationary objects and a stationary model car. However, in future work, we plan to enhance and extend the functionality of the function monitor so that it covers more complex scenarios, with both dynamic and static obstacles. Moreover, we intend to define a method for defining appropriate fail-operational reactions to gracefully degrade the ADS's functionality \cite{Aniculaesei:2019} in response to warning signals given out by the function monitor. Lastly, we plan to integrate the function monitor with the concept of situation monitor, similar to the one presented in \cite{habib2023}. Such integration enables the ADS to be aware of new object classes detected in its environment, thus enhancing its capability to handle novel environment situations. Ultimately, this integration will contribute to the safety and reliability of autonomous driving systems in diverse and challenging real-world scenarios.

%